\documentclass[manuscript,screen]{acmart}

\AtBeginDocument{%
  \providecommand\BibTeX{{%
    \normalfont B\kern-0.5em{\scshape i\kern-0.25em b}\kern-0.8em\TeX}}}

\begin{document}

\title{The X Types--Mapping the Semantics of the Twitter Sphere}

\author{Ogen Schlachet Drukerman}
\affiliation{%
  \institution{University of Haifa}
  \country{Israel}}

\author{Einat Minkov}
\affiliation{%
  \institution{University of Haifa}
  \country{Israel}
}

\renewcommand{\shortauthors}{Drukerman and Minkov}

\begin{abstract}
Social networks form a valuable source of world knowledge, where influential entities correspond to popular accounts. Unlike factual knowledge bases (KBs), which maintain a semantic ontology, structured semantic information is not available on social media. In this work, we consider a social KB of roughly 200K popular Twitter accounts, which denotes entities of interest. We elicit semantic information about those entities. In particular, we associate them with a fine-grained set of 136 semantic types, e.g., determine whether a given entity account belongs to a {\it politician}, or a {\it musical artist}. In the lack of explicit type information in Twitter, we obtain semantic labels for a subset of the accounts via alignment with the KBs of DBpedia and Wikidata. Given the labeled dataset, we finetune a transformer-based text encoder to generate semantic embeddings of the entities based on the contents of their accounts. We then exploit this evidence alongside network-based embeddings to predict the entities' semantic types. In our experiments, we show high type prediction performance on the labeled dataset. Consequently, we apply our type classification model to all of the entity accounts in the social KB.  Our analysis of the results offers insights about the global semantics of the Twitter sphere. We discuss downstream applications that should benefit from semantic type information and the semantic embeddings of social entities generated in this work. In particular, we demonstrate enhanced performance on the key task of entity similarity assessment using this information.
\end{abstract}




\maketitle

\section{Introduction}

Access to popular knowledge is vital for effective information processing and communication by humans and machines alike. To that end, there exist multiple knowledge bases (KBs) that represent factual world knowledge. Alongside Wikipedia,\footnote{https://www.wikipedia.org/} structured knowledge sources like DBpedia\footnote{https://www.DBpedia.org/} and Wikidata,\footnote{https://www.wikidata.org/} describe factual knowledge in a relational form (e.g., as entity-relation triplets, such as  `Obama, {\it served as President of}, the U.S.'). In addition, these resources maintain an ontology of semantic types, associating the entities with those types (e.g., `Obama', {\it is-a}, {\it politician}). It has been shown that the semantic representation of entities in such resources provides useful world knowledge for various text processing applications such as question answering, the disambiguation of named entity mentions, and more~\cite{yamadaTACL17}. A main shortcoming of existing knowledge bases, however, is that they are limited in scope, and are inherently incomplete in their coverage of popular world knowledge~\cite{hoffartCIKM12}. In particular, as of today, public figures such as politicians and musical artists, as well as organizations and businesses, maintain active presence on social networks~\cite{marwick2011see}, where only a minority of those entities are included in existing knowledge bases~\cite{lotan21}. 

In this work, we focus our attention on the social media  service of "X" (previously known as Twitter) as a source of social world knowledge.\footnote{Twitter has officially changed its name to X and URL address to x.com in 2024. As this research was conducted beforehand, we shall use the name Twitter.} Twitter is popular, public, widely studied, and considered as a credible source of knowledge~\cite{castillo2011information}. It is non-trivial however to utilize Twitter as a knowledge base, considering that it is informal, noisy and lacks structured meta-data. Following recent research~\cite{lotan21}, we refer to popular accounts on Twitter, which are followed by large audiences, as {\it entities} of general interest, which form a social KB. While structured KBs associate entities with a semantic ontology of entity types, Twitter lacks structured information about the types of entity accounts. For example, Barack Obama, a former President of the U.S., maintains a highly popular account on Twitter that vaguely describes him as "Dad, husband, President, citizen". In order to exploit social media as a knowledge source, it is desired to infer relevant entity semantics, e.g., identify Obama as a {\it politician}. Concretely, we pose the following main research questions:

\begin{itemize}
\item[RQ1] How, using which evidence, and to what extent is it possible to infer the semantic types of social media entities?
\item[RQ2] What is the distribution of semantic types among popular entities on Twitter?
\item[RQ3] How can one derive meaningful low-dimension embeddings of entities on social media, for the purpose of semantic processing by downstream applications? 
\end{itemize}

We address RQ1 by learning multi-class classification models that perform entity type prediction using evidence that is available within the social network. To overcome the lack of labeled examples, we aligned Twitter accounts with their respective entries in the KBs of DBpedia and Wikidata, where this resulted with the mapping of tens of thousands of popular Twitter account onto a fine-grained schema of 136 semantic types. Using the labeled dataset, we generate a couple of semantic artifacts. First, performing finetuning of the transformer-based neural text encoder of BERT~\cite{bert}, we process textual information that is associated with each Twitter entity into low-dimension embeddings of the entities, that are indicative of their semantic types. We then utilize these semantic text-based entity embeddings alongside network-based embeddings, which have been inferred from a large sample of the Twitter network~\cite{lotan21}, to infer the semantic labels of the entities. Our experiments show high type prediction performance, as evaluated on set aside examples, showing that the best performance is achieved using a combination of the textual and network entity encodings. 

Consequently, we address RQ2 by applying our best performing type prediction model to all of the entity accounts in the social KB--most of which are exclusive to social media. We discuss and validate the results, highlighting the semantics of popular entities in Twitter. To the best of our knowledge, this is a novel effort of mapping the semantics of the `Twitter sphere'. 

In the last part of this work, we address RQ3, showing that the content-based entity embeddings generated in this work capture complementary semantics to network-based information. In a motivating case study, we demonstrate that a low-dimension representation that combines semantic and network aspects of the entities yields preferable results in  assessing social entity similarity. We discuss the implications and potential of using these low-dimension representations in a variety of downstream applications.

This paper proceeds as follows. Following a review of related work (Sec.~\ref{sec:related}), we formulate the entity type prediction task and approach (Sec.~\ref{sec:approach}). We then describe our procedure for aligning Twitter accounts with DBpedia and Wikidata and the resulting labeled dataset (Sec.~\ref{sec:dataset}). The following sections discuss the  semantic entity representations, and our experimental results of type prediction using different combinations of these representations on the labeled dataset~\ref{sec:experiments}. The results of applying the prediction model to all Twitter entities, and a demonstration of semantic entity similarity using our semantic entity representations are given in Sec.~\ref{sec:wild} and~\ref{sec:sim}, respectively. The paper concludes with a summary and a discussion (Sec.~\ref{sec:conclusions}).

\section{Related work}
\label{sec:related}

\paragraph{Semantic entity types in KBs.} In this work, we enrich a social KB of entities with their semantic types. In general, relational factual KBs include entities denoted as nodes and labeled directed links that denote known relations between entity pairs, e.g.,~\cite{laoACL15}. In addition, KBs maintain a hierarchical and transitive semantic backbone, having entities linked to relevant semantic types over special {\it is-a} relations~\cite{dalviWSDM15}. It has been shown that the semantic information associated with entities in a KB is beneficial for a variety of downstream applications.
In question answering, it is common to identify and match a target entity type; e.g., a query that begins with `where' requires an entity of type {\it location} as the answer~\cite{choi2018ultra}. Similarly, information about entity types is useful in applications of entity recommendation or search, where  it is desired to identify relevant entities of a particular type, such as {\it movies}, or {\it actors}~\cite{yuWSDM14}. Further, semantic type information plays a role in the semantic annotation of text, linking entity mentions to the respective KB entity based on a match between the KB entity type and contextual evidence of the entity mention, e.g.,~\cite{yamadaACL16}. Entity type information is also required in relational inference; e.g., a relation of {\it lives-in} requires entities of types {\it person} and {\it location} as agent and object, respectively~\cite{laoACL15}.

\paragraph{Entity type inference.} Traditionally, semantic type inference applied to a handful number of coarse types~\cite{ghaddar2017winer}. More recently, researchers have extended the set of target types. For example, Ling and Weld~\cite{lingAAAI12} presented FIGER, a fine-grained set of 112 semantic classes, which they formed based on popular entity categories in the collaborative database of Freebase~\cite{freebase},\footnote{Now deprecated.}. Aiming to preserve the diversity of Freebase types while removing noise and infrequent types, they manually adapted some types, e.g., unified “dish”, “ingredient”, “food” and “cheese” into a single type named “food”. A later work used FIGER as the target schema in inferring missing entity types in a KB~\cite{yaghoJAIR18}. More recently, researchers proposed to associate entities with free-form types, extracted directly from text, e.g. `skyscraper', `songwriter', or `criminal'~\cite{choi2018ultra}. We explored this possibility in the early stages of this work, considering Twitter account descriptions as the source text. However, we found that account descriptions often lack relevant information. Similar to Ling and Weld~\cite{lingAAAI12}, we consider a schema of fine-grained semantic types, which we induced from the types assigned to a large number of Twitter entities in DBpedia. Our final schema includes 136 types, having excluded types which were rare within the aligned Twitter accounts. 

\paragraph{KB completion} Considering the dynamic nature and vast scope of world knowledge, KBs are inherently incomplete with respect to the covered entities and relations. The task of KB completion applies to the inference of missing information in the KB~\cite{laoACL15,vexler16,neelNAACL15,biswas22}. In particular, researchers previously inferred missing semantic types of KB entities based on contextual evidence collected from the entity mentions in text~\cite{yaghoJAIR18}. Unlike Newswire or Web data, tweets are short, syntactically noisy, and often lack relevant context. In addition, obtaining a sufficient number of entity mentions is computationally costly, and may prove infeasible for many long tail entities. Rather than consider entity mentions, we therefore rely on meta-data, network information, and the historical content associated with each entity account. Our results show that this evidence supports effective semantic type prediction on. Notably, in a closely related work, researchers used Twitter as a source of real-time knowledge for completing missing information in the knowledge base of DBpedia~\cite{nechaevCIKM18}. They performed probabilistic alignment between DBpedia entities and Twitter accounts~\cite{sociallink}, encoded information about each aligned Twitter account as features, and trained an SVM classifier to predict the entity type and other attributes. Similar to our work, they considered text-based encoding of the content posted by each Twitter account, the account's description line, as well as network encodings of popular Twitter accounts that the entity account follows. Compared to our work, the scope of their work was more narrow, focusing on 17 types of interest (including `other'). While their goal was to complete missing information about entities included in a factual KB, we consider Twitter as a social KB, mapping the types of 200K popular accounts onto a comprehensive set of 136 categories. 

\paragraph{Low-dimension entity embedding}
Recently, Lotan and Minkov~\cite{lotan21} outlined SocialVec, a framework for learning network-based entity embeddings in Twitter. Following Word2Vec~\cite{Mikolov2013}, an approach for learning word embeddings from their local contexts, SocialVec models popular entity accounts that individual users co-follow as contextually related. While SocialVec was learned from network evidence only, it was shown to encode semantic aspects, e.g., the accounts of top universities are similar in that they are often co-followed by similar users, where this results in similar network embeddings. On the other hand, network similarity does not necessarily indicate on semantic similarity. For example, the accounts of Obama and Starbucks may be often co-followed by the same users, while belonging to different semantic classes. In this work, we show that content and network information provide complementary evidence with regards to the semantic type of Twitter entity accounts. Specifically, we show that combining SocialVec with content-based entity embeddings yields preferable type prediction and entity similarity assessments,

\section{Problem definition and approach}
\label{sec:approach}

Let us consider a social KB that includes popular Twitter accounts, denoted as {\it entities} $E$. Our goal is to assign each entity $e$ ($e\in E$) with a type $t$ ($t\in T$), where $T$ is a set of target types. For example, $e$ could be Barack Obama and $t$ could be {\it politician}. Similar to related works, we assume a single-label, multi-class classification setting, having each entity assigned to a single type, leaving the case of multi-label classification~\cite{amigACL22} or an incomplete type set~\cite{neelakNAACL15} to future work. This choice is motivated empirically, as we found that the vast majority of Twitter accounts in our labeled dataset was associated with a single type. Our approach of automatic type inference includes the following main sub-tasks:

\paragraph{i. Constructing a labeled dataset} In order to learn models of semantic type prediction, we construct a dataset of example labeled entities. Similar to related works, we take a {\it distant supervision} approach~\cite{lingAAAI12}, exploiting the fact that some Twitter entities have respective entries in structured KBs which are linked with a semantic ontology. This step involves a non-trivial alignment procedure, as well as an adaptation of the KB types to the social media domain. 

\paragraph{ii. Learning models of type prediction.} We use the labeled dataset to learn models of semantic type prediction. Importantly, the models rely on evidence that is intrinsic to Twitter. Hence, the models can be applied to predict the types of the majority of social entities which are not included in factual KBs. More concretely, we consider various types of evidence, pertaining to meta-data such as the account's description line, the social network,  and the historical tweets posted by each account. A main challenge concerns the effective processing and integration of incompatible textual and network-based information in learning. We propose to first finetune account-level textual embeddings using a transformer-based contextual text encoder, where we then combine this evidence with network-based user embeddings using a dedicated neural network. 

\paragraph{iii. Applying type inference at large-scale.} following tuning experiments using the labeled dataset, We apply the best performing model to predict the semantic types of the non-aligned entities within the social KB. The analysis of the resulting type assignments suggests insights about the global distribution of semantic types among popular accounts in Twitter. 

Overall, this research yields two main artifacts: the assignment of semantic types to Twitter entities, and low-dimension text-based embeddings of those entities as tuned on the semantic type prediction task. 

\section{Dataset construction via alignment to DBpedia and Wikidata}
\label{sec:dataset}

Well-known personalities and organizations who maintain active accounts on social media are likely to be indexed in factual KBs. We wish to align Twitter entities with their entries in existing KBs, which associate entities with semantic ontologies~\cite{dalviWSDM15}, so as to label the aligned entity accounts according to this semantic type information. We focus our attention on Wikidata and DBpedia as reference KBs. {\it Wikidata}\footnote{https://www.wikidata.org/}~\cite{pellissierWWW16} is a popular collaborative knowledge graph developed and maintained by the Wikimedia Foundation, which is best known for hosting the Wikipedia project. Wikidata serves as a multilingual, centralized, linked data repository for all of Wikimedia projects. It forms a very large graph that includes more than a hundred million entities. Wikidata represents taxonomic hierarchical semantic relations ('instance of'), inter-entity relationships, and entity properties, providing high-quality world knowledge of wide coverage~\cite{farber2018knowledge, farber2015comparative,shenoy2022study}. {\it DBpedia} is another collaborative project which maps structured information from Wikipedia (mainly, from 'infobox' tables and other templates found in most Wikipedia pages) onto a semantic ontology~\cite{DBpedia}. We found the ontology maintained by Wikidata to be somewhat intricate and abstract for our purposes, and favoring the ontology by DBpedia, which defines a hierarchical tree-like structure of semantic types. Nevertheless, the two sources are complementary. Importantly, Wikidata often specifies the Twitter account ID of entities. We exploit this information for aligning Twitter accounts against Wikidata, which is in turn linked to DBpedia. In addition, Wikidata is of wider coverage. We utilize textual entity descriptions that are available in Wikidata to refine and extend our labeled dataset. This section describes the entity alignment and labeling procedure, and the resulting labeled dataset.

\subsection{The alignment procedure} 

Aligning Twitter accounts with their respective entries in Wikidata and DBpedia is a non-trivial task. In a recent effort, researchers performed probabilistic alignment of Twitter accounts against DBpedia~\cite{sociallink}, modeling graph-based features on both sides, and using a deep learning model to select the best matching candidate for each target entity. Their results indicated on a non-negligible ratio of alignment errors. In our work, we opt for a deterministic  approach, which indicates on alignments at nearly perfect accuracy, where our goal is to construct a high-quality labeled dataset.

In brief, there are two steps of our alignment procedure. First, we exploit the fact that Wikidata often specifies the Twitter account ID of a given entity.\footnote{Item property: `social media followers'}. Since it is not possible to retrieve a Wikidata entry given a Twitter account ID using a public API, we perform exhaustive reverse search, seeking matches against the Twitter account identifiers in our social KB. Thus, the alignment against Wikidata is deterministic and precise, to the extent that the information in Wikidata is accurate. Unlike Wikidata, DBpedia does not directly associate entities with their Twitter identifiers. DBpedia links each entity to Wikidata however via the Wikidata Qid field. To align the accounts with DBpedia, we therefore first identify candidate entity pages in DBpedia, having the Twitter account name submitted as a search query to DBpedia~\cite{mendes2011evaluating,gillick2014context}. Then, scanning the candidates, we verify a match based on the Wikidata Qid field information. While additional DBpedia pages may be mapped to Twitter accounts probabilistically, e.g., based on string matching, we abstain from incorporating uncertainty into the alignment process.

\begin{figure}[t]
\centering
\includegraphics[width=0.5\columnwidth]{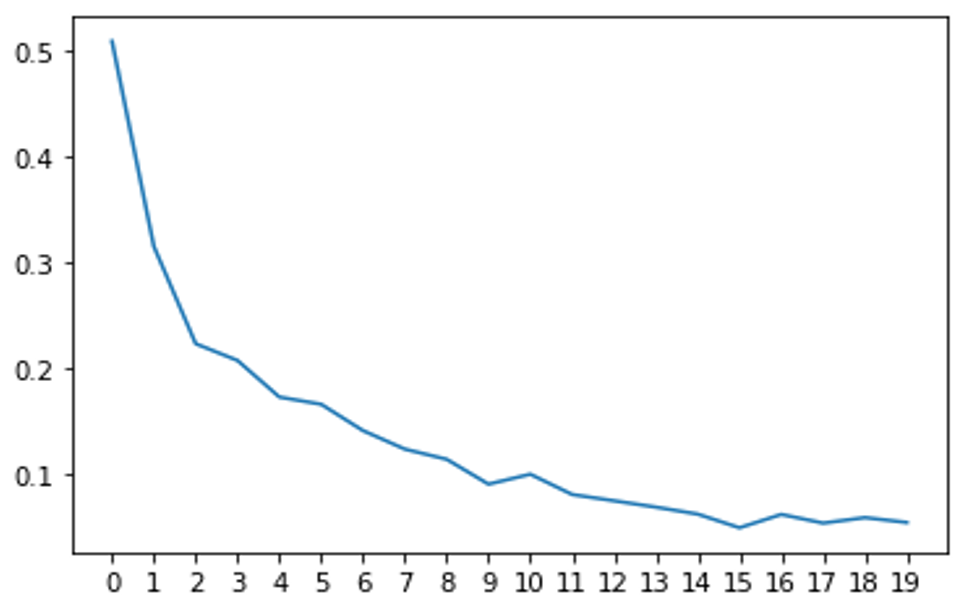}
\small
\caption{The ratio of Twitter entity accounts which we successfully aligned with DBpedia, listed per 10K accounts bins that are ordered by descending popularity (where bin \#0 includes the most popular 10K accounts). As shown, the ratio of accounts that align with entries in DBpedia correlates with the account popularity in Twitter, that is, popular Twitter entities are more likely to be covered by DBpedia.}
\label{fig:linkage-ratio}
\end{figure}

Considering all Twitter accounts in our social KB (200K), we successfully matched Wikidata items for 43K accounts (21.5\%), out of which 26.5K (13.3\%) were also successfully mapped onto DBpedia (mapping in itself does not guarantee an available semantic type). Presumably, popular accounts, as measured by the number of their followers, are more likely to be included in factual KBs. Figure~\ref{fig:linkage-ratio} confirms this correlation, showing decreasing alignment rates by account popularity. As shown, about 50\% of the 10K most popular entities (top 5\% popularity-wise) were successfully aligned with DBpedia, as opposed to less than 10\% of the least popular 10K accounts. 

\subsection{The labeling schema}

\begin{figure}[h]
\centering
\includegraphics[width=0.95\columnwidth]{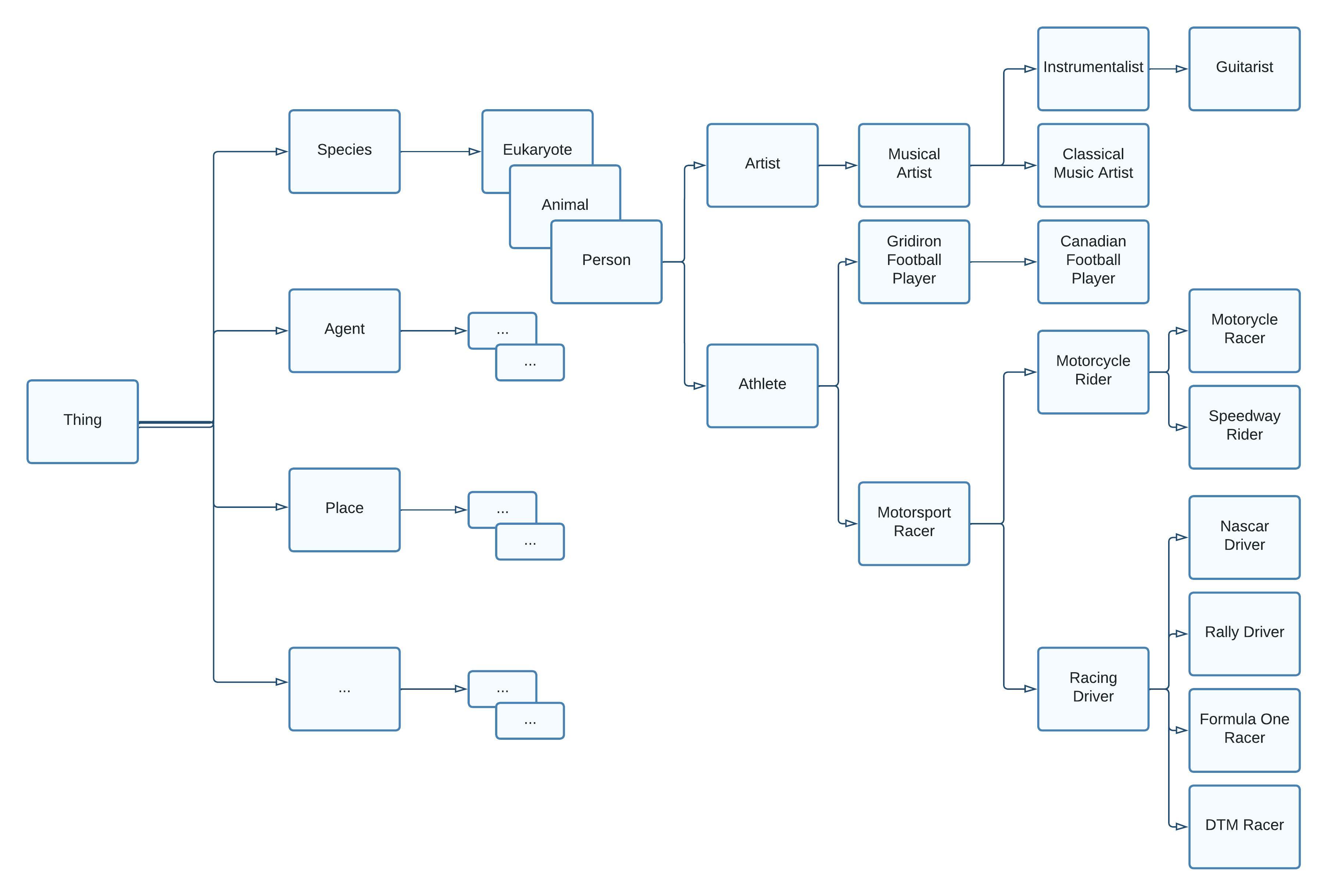}
\small
\caption{Illustration of selected paths extracted from DBpedia's hierarchical structure of semantic types}
\label{fig:DBpedia-paths}
\end{figure}

For those entity accounts aligned with DBpedia, we wish to obtain their semantic types from this resource. Notably, DBpedia's ontology comprises a deep hierarchical of a coarse-to-fine structure of roughly 800 semantic categories~\cite{pattuelli2013knowledge}.~\footnote{http://mappings.DBpedia.org/server/ontology/} Figure~\ref{fig:DBpedia-paths} illustrates several paths from this hierarchy; for example, it shows that the category of `Guitarist' is reached over the following 8-hop path, starting from the root category of 'Thing': {\it Thing} $\rightarrow$
{\it Species} $\rightarrow$ {\it Eukaryote} $\rightarrow$ {\it Animal} $\rightarrow$ {\it Person} $\rightarrow$ {\it Artist} $\rightarrow$ {\it Musical artist} $\rightarrow$ {\it Instrumentalist} $\rightarrow$ {\it Guitarist}. We found that leaf nodes reside at different levels of the hierarchy tree, yet the the majority of the most specialized types, i.e., leaf nodes, correspond to paths of length six or less. Considering that the top three levels are abstract, we consider the semantic categories that reside at level 4 or below in DBpedia's hierarchy. 

\begin{figure}[h]
\centering
\includegraphics[width=0.95\columnwidth]{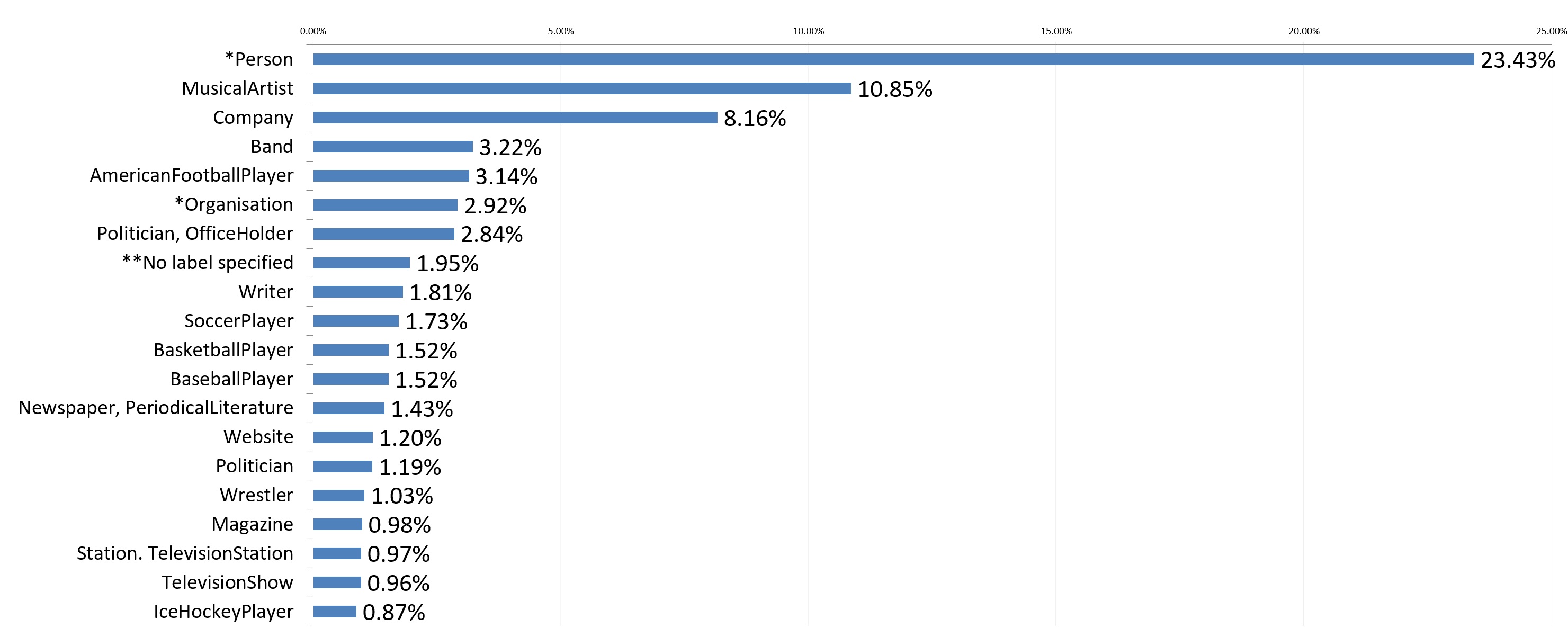}
\small
\caption{The 20 most frequent semantic paths among the DBpedia pages that map to popular Twitter entities. Overall, these paths apply to 71.7\% of all aligned Twitter entities.}
\label{fig:top-20-paths}
\end{figure}

Figure~\ref{fig:top-20-paths} lists the most prevalent semantic paths\footnote{We exclude the top three levels of the hierarchy.} among the aligned Twitter entities. Overall, there are 1,002 distinct paths associated with those entities, which demonstrate a 'long tail' distribution. The 20 most prevalent paths, which are displayed in Figure~\ref{fig:top-20-paths}, account for almost 70\% of the label assignments. As shown, the assigned types are of varying granularity. The most common type is the general {\it person} type. The following most common semantic paths are {\it person.artist.Musical Artist} and {\it organization.Company}. 

In performing entity type prediction, we will consider the high-level semantic categories of {\it person}, {\it organization}, {\it place}, {\it work} and {\it other} as coarse entity type. Further, we aim to make informative distinction between fine-grained entity types~\cite{lingAAAI12,choi2018ultra}. We make the design choice of predicting the most specialized types that the entities are associated with, i.e., the leaf nodes of the semantic hierarchy.  (In our experiments, we examined also the direct descendants of the entity's coarse semantic type as the target schema (e.g., {\it artist} as opposed to {\it musical artist}), however, prediction performance was roughly the same at both resolutions. We therefore favored the finer distinction.) Removing rare types with less than 5 occurrences in our dataset eliminated about 40\% of the types, which accounted for less than one percent of the labeled examples. 

\begin{table}[t]
\small
\centering
\begin{tabular}{lll}
\hline
\textbf{PERSON}	&	Presenter - Radio Host	&	Artwork	\\
Actor	&	Producer	&	Book	\\
Artist	&	Scientist	&	Film	\\
. - Adult Actor	&	Sports Manager	&	Magazine	\\
. - Comedian	&	Surfer	&	Musical Single	\\
. - Comics Creator	&	Television Host	&	Newspaper	\\
. - Fashion Designer	&	Voice Actor	&	Periodical Literature	\\
. - Musical Artist	&	Wrestler	&	Radio Program	\\
Astronaut	&	Writer	&	Software	\\
Athlete	&	Youtuber	&	Television Show	\\
. - American Football Player	&		&	Video Game	\\
. - Baseball Player	&	\textbf{ORGANIZATION}	&	Website	\\
. - Basketball Player	&	Band	&		\\
. - Cricketer	&	Broadcaster	&	\textbf{PLACE}	\\
. - Cyclist	&	. - Broadcast Network	&	Country	\\
. - Figure Skater	&	. - Radio Station	&	Settlement	\\
. - Golf Player	&	Company	&		\\
. - Gymnast	&	. - Airline	&	\textbf{OTHER}	\\
. - IceHockey Player	&	. - Bank	&	Airport	\\
. - Martial Artist	&	. - Brewery	&	Architectural Structure	\\
. - Motorcycle Rider	&	.  - LawFirm	&	Automobile	\\
. - Nascar Driver	&	. - Public Transit System	&	Award	\\
. - Poker Player	&	. - Publisher	&	Basketball League	\\
. - Racing Driver	&	Educational Institution - Library	&	Beverage	\\
. - Skier	&	Government Agency	&	Building	\\
. - Soccer Player	&	Legislature	&	Convention	\\
. - Swimmer	&	Military Unit	&	Fictional Character	\\
. - Tennis Player	&	Political Party	&	Film Festival	\\
Beauty Queen	&	Record Label	&	Food	\\
Boxer	&	School	&	Golf Tournament	\\
Business Person	&	Soccer League	&	Hospital	\\
Chef	&	Sports Club - Rugby Club	&	Hotel	\\
Chess Player	&	Sports Club - Soccer Club	&	Information Appliance	\\
Cleric	&	Sports League	&	Military Conflict	\\
College Coach	&	. - Ice Hockey League	&	Museum	\\
Criminal	&	Sports Team	&	Music Genre	\\
Dancer	&	. - American Football Team	&	Person Function	\\
Economist	&	. - Baseball Team	&	Restaurant	\\
FormulaOne Racer	&	. - BasketballTeam	&	Soccer Tournament	\\
Journalist	&	. - CricketTeam	&	Societal Event	\\
Lawyer	&	. - HockeyTeam	&	Space Mission	\\
Military Person	&	TelevisionStation	&	Sports Event	\\
Model	&	Trade Union	&	Stadium	\\
Movie Director	&	University	&	Station	\\
Philosopher	&		&	Theatre	\\
Photographer	&	\textbf{WORK}	&	Time Period	\\
Playboy Playmate	&	Album	&	Venue	\\
Politician	&	Academic Journal	&		\\
\hline
\end{tabular}
\caption{The target categories}
\label{tab:schema}
\end{table}

The final labeling schema is presented in Table~\ref{tab:schema}. Overall, it includes 5 coarse categories (including `other'), and 136 fine-grained categories. Each entity in the dataset is associated with its fine-grained type, as well as with the ancestor coarse type; e.g., {\it musical artist} and {\it person}. While bearing a similarity to related entity labeling schemes~\cite{lingAAAI12}, the target types defined in this work originate from the type distribution of popular Twitter accounts. 

\subsection{The dataset}

\begin{table}[t]
\small
\centering
\begin{tabular}{llr}
\hline	
KB source & Label source & Size \\
\hline
DBpedia & Fine-grained labels available &    23,800\\
Wikidata & Fine-grained labels induced via weak supervision &    5,600\\
DBpedia \& Wikidata & Coarse DBpedia labels specialized into fine-grained labels using weak supervision &    5,000\\
\hline	
Total &   & 34,400\\
\hline	
\end{tabular}
\caption{Dataset statistics: the number of entities obtained via alignment to each reference KB, either directly or by means of weak supervision.}
\label{tab:meta_data_summary}
\end{table}

\begin{table}[t]
\small
\centering
\begin{tabular}{llrr}
Coarse & Fine & Ratio [\%] & Cumulative [\%]\\
\hline
Org	&	Company	&	0.14	&	0.14	\\
Person	&	Actor	&	0.12	&	0.26	\\
Person	&	Musical Artist	&	0.12	&	0.38	\\
Person	&	Politician	&	0.05	&	0.43	\\
Person	&	Journalist	&	0.04	&	0.47	\\
Person	&	Writer	&	0.04	&	0.51	\\
Org	&	Band	&	0.03	&	0.54	\\
Person	&	American Football Player	&	0.03	&	0.57	\\
Work	&	Website	&	0.02	&	0.59	\\
Org	&	Television Station	&	0.02	&	0.61	\\
Person	&	Soccer Player	&	0.02	&	0.63	\\
Other	&	Station	&	0.02	&	0.64	\\
Work	&	Newspaper	&	0.02	&	0.66	\\
Person	&	Comedian	&	0.02	&	0.67	\\
Person	&	Basketball Player	&	0.01	&	0.69	\\
Person	&	Baseball Player	&	0.01	&	0.70	\\
Work	&	Magazine	&	0.01	&	0.71	\\
Person	&	Wrestler	&	0.01	&	0.73	\\
Person	&	Model	&	0.01	&	0.74	\\
Work	&	Television Show	&	0.01	&	0.75	\\
Person	&	Scientist	&	0.01	&	0.76	\\
Org	&	Radio Station	&	0.01	&	0.77	\\
Person	&	Business Person	&	0.01	&	0.78	\\
Person	&	Ice Hockey Player	&	0.01	&	0.79	\\
Org	&	Soccer Club	&	0.01	&	0.80	\\
\hline
\end{tabular}
\caption{The top 25 final semantic types assigned to labeled Twitter accounts of entities based on alignment to Wikidata and DBpedia and augmentation via weak supervision. We use this dataset for training and evaluating classifiers of type prediction.}
\label{tab:dataset}
\end{table}

We found that for many entities, a fine-grained category was not fully specified in DBpedia. For example, as shown in Figure~\ref{fig:DBpedia-paths}, the coarse category of {\it person} was assigned to 23\% of the aligned entities, with no further specification. Similarly, 3\% of the entities were merely labeled as {\it organization}. Moreover, no semantic label whatsoever was available for 2\% of the aligned entities. In order to complement the missing type information, we referred to the textual descriptions about each entity in Wikidata, which are typically suggestive of their semantic type; for example, the description of entities of type {\it actor} is likely to includes the words `actor' or `actress'. To that end, we trained a classifier to determine the semantic labels of entities from their textual descriptions in Wikidata. We used the aligned accounts with fully specified types (23.8K) as training examples, finetuning a BERT-base-uncased architecture on the type prediction multi-class classification task. The classifier's performance, evaluated on a held out stratified sample of 1\% of the labeled data, measured 0.86 in terms of weighted F1. While imperfect, inferring fine-grained labels in this fashion allowed us to extend that dataset with additional 10.6K accounts that have been aligned with Wikidata but not with DBpedia. Table~\ref{tab:meta_data_summary} summarizes the statistics of our labeled dataset. Overall, type labeling via weak supervision increased the labeled example set size by about 45\%, amounting to 34.4K labeled entities in total. Table~\ref{tab:dataset} lists the most prevalent semantic types of the entities in the final dataset. As shown, the 25 most frequent semantic categories account for 80\% of the labeled instances. 

\section{Type prediction: Model and experiments}
\label{sec:experiments}

Our aim is to automatically infer semantic types of unaligned Twitter accounts based on public information that is available about those accounts within the social network. In this section, we first outline multiple entity representation schemes, which describe content- and network-based evidence about each entity, and then describe our learning approach and experimental results using the labeled dataset. 

\subsection{Entity representations}

As described below, we form and gauge multi-facet entity representations. We obtain relevant evidence in an efficient and scalable fashion. The various evidence is then processed into user-level low-dimension embeddings.

\paragraph{Historical tweets.} The tweets posted by a social media user provide valuable evidence about its type. Consider the following excerpts extracted from Barack Obama's posts: "As chancellor, Angela Merkel served with integrity.."; or, "Senator Bob Dole was a war hero, a political leader and a stateman". This content suggests that the account belongs to a {\it politician}. We obtained up to 100 recent tweets for each entity via Twitter API. Given 200K entities, this resulted in a corpus of roughly 20M tweets, which we collected over several months in 2022. Since the tweets form a sequence of non-consecutive texts, we generate an encoding of each tweet separately, and represent the user as the average vector of their tweet encodings. Formally, let us assume that there are $N_e$ historical tweets available for entity account $e$. We denote the vector encodings of the individual tweets as $v_i^e$, $i=1,..,N_e$, and the encoding of the entity as $e^T={\sum_{i=1}^{N_e} v_i^e \over {N_e}}$. Specifically, we generate the text encodings using BERT~\cite{bert}, a neural pretrained transformer-based text encoder, which has been  shown to yield high-quality results on tweets. Considering that tweets may be noisy and incoherent, we also consider representing the tweets as simple average of non-contextual word embeddings~\cite{gloveEMNLP14}. 

\paragraph{User descriptions.} While filling a self-description is voluntary in Twitter, this information is available for more than 90\% of the users in our social KB. Presumably, some textual descriptions would include indications regarding the semantic type of the account, e.g., "astrophysicist". In practice, account owners often fill this field with other statements; for example, the description of the singer Katy Perry was "Love. Light.". Consider the description line as potential evidence, we process it using the text encoder of BERT into a low-dimension vector, $e^D$.

\paragraph{Network information.}
Social media users associate themselves with, aka {\it follow}, other accounts of interest,  over structured links. One's network profile may therefore provide valuable information about them. In this work, we represent network information using the pre-trained SocialVec embeddings~\cite{lotan21}. SocialVec embeddings have been learned from a large sample of $\sim$1.5M random users of Twitter, aiming to capture similarity between popular entities that are co-followed by the same users. In addition to modeling social similarity, the resulting entity embedding vectors have been shown to encode information related to semantic types. Indeed, entities in some given domain, e.g., sports, may be followed by similar users, and thus reside close to each other in the social embedding space. In our experiments, we thus represent each entity using its network-based SocialVec embedding, $e^{N}$, assessing the extent to which this representation supports the prediction of the entity's fine-grained semantic type.  

\paragraph{Multi-facet entity representations}

The various entity embeddings--which describe its historical tweets ($e^T$), description ($e^D$) and network information ($e^N$)--may be combined via concatenation to produce a multi-facet user representation. In our experiments, we investigate the utility of each representation scheme independently, and in combination, and determine the representation scheme which yields the best performance on the semantic type prediction task.

\subsection{Learning approach}

\begin{figure}[t]
\centering
\includegraphics[width=0.85\columnwidth]{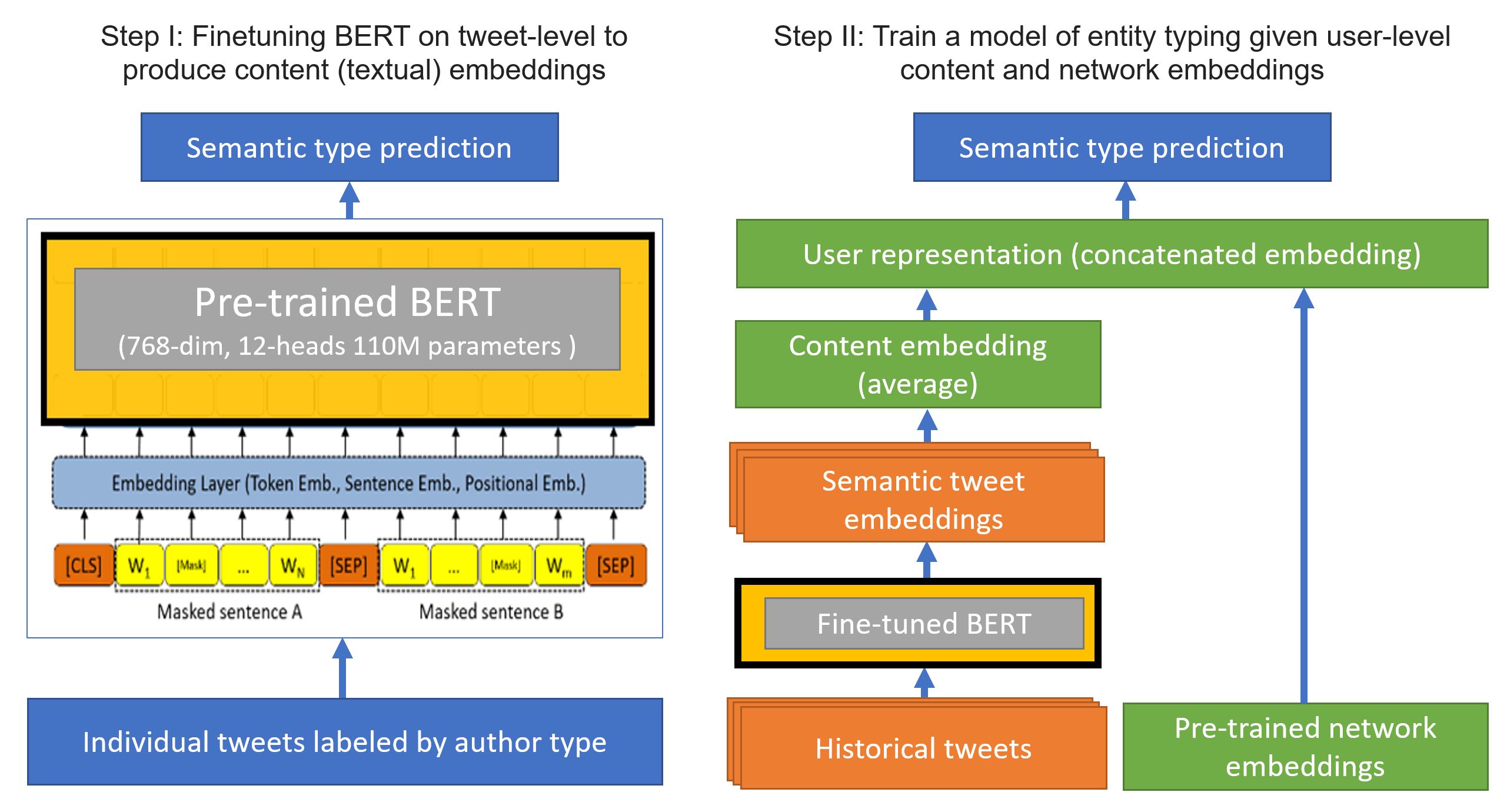}
\small
\caption{A multi-step learning approach: First, we fine-tune BERT on individual tweets from our training set, attributing the account label to each tweet. The embeddings of tweets using the finetuned model are then aggregated (averaged) at user level. Similarly, social embeddings of popular entity accounts followed by the user, which were learned from a large sample of the Twitter network, are also aggregated (averaged) to form a social network-based encoding of the user. The content- and social-based embeddings may be then concatenated and fed to a neural network, trained to predict the accounts' semantic label based on this multi-facet evidence.}
\label{fig:method}
\end{figure}

Figure~\ref{fig:method} illustrates the main step of our approach, learning to predict entity types given the various entity representations: 
\begin{enumerate}
\item {\it Finetuning BERT.} As a first step, we finetune BERT on the task of semantic typing, predicting the type of each labeled entity in the training set given their historical tweets. Notably, the historical tweets by a given entity must be processed at tweet level rather than user level, as transformer-based encoders are limited with respect to input length~\cite{bert}. We therefore predict the semantic label of an account given each of their individual tweets, where we practically assign each tweet the type of the account that authored it. While some tweets may not provide type-related evidence, we make the assumption that there generally exists relevant evidence at tweet level. This step is depicted in the left part of Figure~\ref{fig:method}. In finetuning BERT, we follow the common practice of extending its output text encoding with a terminal feed-forward network~\cite{bert}, training the entire network to predict the given text label. The {\it finetuning} process specializes BERT on the task of semantic type prediction, so that the text encodings generated by the finetuned model highlight relevant semantic aspects.  

\item {\it Constructing the content-based entity embeddings.} As shown in the right part of Figure~\ref{fig:method}, following the finetuning step, we apply the specialized BERT model to produce semantic encodings (the output CLS vector of the finetuned model) of each tweet by a given entity account. The content-based embedding of the entity is then computed as the average of its tweet encodings. We have also experimented with aggregating the tweet embeddings using max and min operations, as well as with the possibility of majority voting, but this has led to inferior performance in our experiments. While the figure does not refer to the description line as an information source, it is processed in a similar fashion, excluding the average operation.

\item {\it Type prediction.} Finally, the resulting content-based vectors of the entity and its pre-trained SocialVec network-based embedding are con-joined via vector concatenation. The full vector is input to a dedicated neural network classifier with a softmax terminal layer over the target semantic classes. In the experiments, we found it useful to employ a weighted loss function L:
\begin{equation}
L = L_1\cdot\alpha + L_2\cdot\beta + L_3\cdot\gamma
\label{eq:exp}
\end{equation} 
where $L_{1}$ is the loss considering only the input vector coordinates that map to the Socialvec embedding, $L_{2}$ is the loss due to the content-based embedding coordinates, and $L_{3}$ is a non-restricted standard loss function. Such a mixed cost function encourages the model to attribute importance to both types of evidence rather than focus on a single encoding type, which may be dominant with respect to input length, or more correlated with the label class. 
\end{enumerate}

\subsection{Experimental setup}

In our experiments, we employ BERT using its public implementation.\footnote{https://huggingface.co/google-bert/bert-base-uncased}. We also experimented with BERT-large and ROBERTA~\cite{roberta}, however these larger models resulted in comparable performance. The finetuning of BERT on the task of type prediction required 4 epochs, optimizing performance on a held-out validation set. Overall, the finetuning step was performed on a single GPU (Nvidia A100) for a duration of 8 hours. Following finetuning, we obtained the CLS 768-dimension encodings of each tweet. We also employed word-level embeddings as alternative to the contextual BERT, using the GLOVE word embeddings~\cite{gloveEMNLP14}. The GLOVE model provides 200-dimension embeddings for 1.2M distinct tokens, learned from 2 billion tweets..\footnote{https://nlp.stanford.edu/projects/glove/} In this case, an entity is presented by averaging the GLOVE embeddings of all the tokens contained in its historical tweets. Finally, the SocialVec entity embeddings are of 100 dimensions, obtained from a public repository~\cite{lotan21}.

The various entity representation vectors are input to a dedicated classifier, which is trained to predict the semantic labels using the train set. While we attempted classification using a variety of methods, including XGBoost, the memory-based KNN, and SVM, we found that logistic regression and a multi-layer fully connected neural network yielded the best performance. We report our results using the neural classifier. Based on the results of tuning experiments evaluated on a held out validation set, we configured the neural network classifier to 3-layers, including a single intermediate layer of 50 nodes. Having performed grid search within the range 1-10, We set the weight parameter values in Eq.~\ref{eq:exp} to $\alpha=5$, $\beta=1$ and $\gamma=1$. Overall, we found that weighted cost function improved classification performance by absolute 2.8\% in weighted F1, compared to a standard loss function ($L_3$ only). Hence, we report our results using the enhanced loss function.

\subsection{Type classification results}

\begin{table}[t]
\small
\centering
\begin{tabular}{ll|ccc|ccc}
& Text & \multicolumn{3}{c}{Gold-labeled train set}	& \multicolumn{3}{|c}{Extended train set}	\\
Information source & embeddings & Weighted F1 & Macro-F1 & Accuracy & Weighted F1 & Macro-F1 & Accuracy \\
\hline	
\multicolumn{8}{l}{\textbf{SINGLE-SOURCE}} \\ 
\hline
	Content (tweets) &	GloVe 	&	0.559	& 0.316 & 0.592	&	0.545 & 0.335 & 0.576 \\
	&	BERT 	&	0.388	& 0.082 & 0.480 &	0.463	& 0.148 & 0.532 \\    
        &	BERT FT 	&	\textbf{0.638}	& \textbf{0.411} & \textbf{0.656} &	\textbf{0.647}	& \textbf{0.428} & \textbf{0.660} \\
        \hline
Network	&	-	&	0.557	& 0.321 & 0.568 & 	0.556 & 0.320 & 0.554 \\
\hline
Description	&	BERT FT	&	0.445 & 0.136 & 0.522 	&	0.499 & 0.179 & 0.548 	\\

\hline
\multicolumn{8}{l}{\textbf{MULTI-SOURCE}} \\ 
\hline
Network \& content	& Glove	&	0.605	& 0.353 & 0.633 & 0.608	& 0.373 & 0.631 \\
	& BERT	&	0.633	& 0.389 & 0.663 &	0.634	& 0.391 & 0.650 \\
	& BERT FT	&	\textbf{0.683}	& \textbf{0.473} & \textbf{0.694} &	\textbf{0.683}	& \textbf{0.458} & \textbf{0.690} \\
Network, content \& desc.	& BERT	FT	&	0.653 & 0.451 & 0.670 &	0.672 & 0.441 & 0.671	\\
\hline
\end{tabular}
\caption{Classification results as measured on gold labeled set-aside test examples, for which the semantic types were determined based on alignment to DBpedia. There are 136 target types overall. The table presents results using gold-labeled train examples vs. an extended train set which includes additional weakly labeled examples. Multiple evidence sources are used, either independently or in combination. The best classifier uses network and content evidence.}
\label{tab:results}
\end{table}

Table~\ref{tab:results} includes our main experimental results of fine-grained type classification using the various entity encodings as evidence. We evaluate the entity typing classifiers on a randomly sampled test set of 3.8K ($\sim$15\%) of the labeled accounts. The types of all of the test examples were assigned via alignment to DBpedia, serving as a gold-labeled test set in evaluating classification performance. The results are reported in terms of Macro-F1, averaging the F1 scores over all of the target classes, and Weighted F1, having the F1 scores weighted by type frequency. In addition, we report accuracy, defined as the ratio of test examples for which the classifier prediction matched precisely the gold label type. As classification performance is generally better for prevalent categories, for which there exist ample training instances, the reported weighted F1 and accuracy figures are higher compared to macro-F1, which attributes equal importance to rare types. Below, we refer to weighted F1 as the main evaluation metric. 

Table~\ref{tab:results} details the results of two sets of experiments, training the models using either the `gold labeled' accounts, which we successfully aligned with DBpedia, or using the `extended' dataset obtained via weak supervision (see Table~\ref{tab:dataset}). In both setups, we evaluate performance on the same test entities. Overall, we observe better or comparable performance using the extended dataset. For example, type prediction based on the content encodings (historical tweets) measures 0.46 versus 0.39 in terms weighted F1. However, the best typing results in both setups are comparable, measuring 0.69 in weighted F1 using both setups. We believe that incorporating network entity embeddings provides complementary evidence, diminishing the impact of the additional training examples in the extended dataset. In general, we consider the extended dataset to be a valuable resource for further studying type prediction and related problems, and make this resource available to the research community.

Next, let us review the results by information source. As shown, using the content (tweets) posted by the account as sole evidence yields varying results, depending on the content representation scheme. Encoding the tweets in terms of Glove word embeddings gives F1 performance of 0.56.\footnote{For simplicity, we refer to the results obtained using the gold-labeled dataset. Similar trends are observed for the extended dataset.} The contextual encodings of the tweets generated using the pretrained BERT (`BERT') yield substantially lower performance, measuring 0.39 in F1. Nevertheless, encoding the tweets using the finetuned BERT model (`BERT FT') yields the best performance, reaching F1 of 0.64. Modeling network-based information as standalone evidence yields second-best F1 performance of 0.56, showing that network information is suggestive of semantic types. Finally, modeling the description line as sole evidence shows that it carries less information compared with the other sources, reaching F1 performance of 0.45. The bottom part of Table~\ref{tab:results} shows classification performance given different combinations of the entity embedding vectors, joined via concatenation. We find that combining network and content evidence gives the best result overall, measuring 0.69 in weighted F1. Next, we will employ this best-performing classification configuration towards semantic type prediction for all of the Twitter accounts in our social KB. Table~\ref{tab:coarse} details the results of the selected classifier on the same test set at coarse type resolution. As shown, F1 performance is as high as 0.96 on the {\it person} category, which is the most prevalent in the dataset. Overall, coarse type performance measures 0.84 in terms of weighted F1, compared with weighted F1 performance of 0.69 in predicting the entity types at fine resolution. 

\begin{table}[t]
\small
\centering
\begin{tabular}{lcccr}
 	&	Precision	&	Recall	&	F1	& \# Examples	\\
  \hline
Person	&	0.945	&	0.967	&	0.956	&	2572	\\
Organisation	&	0.747	&	0.739	&	0.743	&	1164	\\
Work	&	0.678	&	0.579	&	0.625	&	437	\\
Other	&	0.375	&	0.426	&	0.399	&	176	\\
Place	&	0.526	&	0.385	&	0.444	&	26	\\
\hline
\end{tabular}
\caption{Classification results using our best performing model evaluated at coarse resolution}
\label{tab:coarse}
\end{table}

\section{Entity typing `in the wild'}
\label{sec:wild}

There remain 157K (out of 200K) popular Twitter accounts that could not be aligned with the reference KBs. The semantics of those entities must therefore be inferred automatically. In this section, we first examine the generalization of our semantic typing model to a random sample of Twitter entities. We then describe the outcomes of processing entity semantics at large scale, and our insights regarding the semantics of the Twitter sphere. 

\subsection{Evaluation of classifier generalization}

There may exist a distribution shift between the aligned entities and the other Twitter accounts. To examine the generalization of the automatic typing model, we randomly sampled 240 non-aligned entity accounts, and manually examined their types. (Placing more emphasis on the most popular accounts, we first split the population of entity accounts into 5 bins based on their number of followers. We then sampled 120 accounts from the top bin, and further sampled 30 accounts from each of the remaining bins.) In a manual annotation effort, the first author examined relevant public information about the selected accounts within Twitter or the Web, aiming to assign a semantic type to each entity out of our target schema of 136 types. In some cases, the description and content posted by the account in Twitter provided inconclusive evidence, making it hard to reliably determine the semantic type of the entity account. Consequently, the manual annotation resulted in the assignment of semantic labels to 199 entities. In order to alleviate ambiguity and subjectivity, the annotation process accommodated the assignment of a secondary label for a given entity, e.g., a  {\it Writer} who is also a {\it Producer}; or, a {\it Magazine} that may be also considered as a {\it Newspaper}. A secondary type was assigned for 65\% of the labeled entities. 

We note that during annotation, we also captured semantic types in free-form, noting semantic types that were not available in the target semantic scheme. The free form types mainly pertained to {\it social interest groups}, {\it digital news}, {\it public mentors}, {\it content creators}, {\it entrepreneurs} and {\it business executives}. The fact that these categories are missing from our labeled dataset, and hence from the target type schema, is a limitation of this research and a venue for future research.

We applied and evaluated the performance of the type prediction models trained using the aligned accounts with respect to the manually labeled sample. Since the sample is limited in size, we assess the results in terms of accuracy, measured as the ratio of examples for which the predicted type matched the manually assigned label (primary or secondary). Considering models trained using different information sources, we observed similar trends to our tuning experiments (Table~\ref{tab:results}): the best performance was achieved using a combination of network- and content-based evidence. Classification accuracy on the manually annotated sample was substantially lower however, measuring 0.30 compared to 0.69 on the aligned and gold labeled examples. 

\paragraph{Error analysis}

\begin{table}[t]
\small
\centering
\begin{tabular}{llll}
\hline	
Mismatch due to: & Predicted type & Manually assigned type & Topics of tweets \\
\hline
\multicolumn{4}{l}{\it Subjectivity of the annotation} \\
\hline
Semantic ambiguity  & Magazine & Website & Health \\
(10/25) & Company & Website & Travel; organic foods; education \& technology \\
    & Writer & Business person & Business  mentoring; investments \\
 & Record label & Company & Music events\\
\hline	
Multi-label entities & Writer&Musical artist&Unclear\\
(5/25) & Politician&Writer&Politics\\
 & Company&Writer&Commercial\\
 & Model&Writer&Mentoring\\
 & Basketball player & Business person & Sports, real estate\\
\hline	
\multicolumn{4}{l}{\it Classification errors} \\
\hline
 Incorrect yet topically & Business person & Website & Capital markets\\
 meaningful  & Journalist & Website & Liverpool FC, sports \\
(7/25) & Musical artist & Website & Unclear \\
 & Company & Writer & Business, mentoring\\
 & Baseball player & Website & Unclear, sports\\
 & Musical artist & Company & Music \\
 & Company&Business person& Health, business/marketing\\
\hline	
Misleading evidence & Basketball player& Musical artist&Unclear, sports \\
(3/25) & Politician & Musical artist & Politics\\
 & Band & Company & Unclear, alcohol \\
\hline
\end{tabular}
\caption{Error analysis of 25 type prediction mistakes, detailing only the uniquely predicted labels versus the manually assigned labels, and the main topics observed within the account's posts.}
\label{tab:unlabeled_error_analysis}
\end{table}

In order to analyze the reasons for the low classification accuracy on our annotated sample, we performed error analysis for 25 random entities with erroneous predictions. Table~\ref{tab:unlabeled_error_analysis} lists the reasons for the mismatching type assignments, grouping the examples by assigned versus predicted types, and detailing the main topics observed for the accounts. As shown, in 40\% of the cases (10/25), we found the predicted type of the the entities to be sensible, highlighting relevant rather than wrong semantic aspects of the entity; e.g., some entities labeled as {\it website} were predicted to be a {\it company} or {\it magazine}; a {\it business person} was classified as a {\it writer}, and a {\it company} was predicted as a {\it record label}. In all of these cases, the annotator found the predicted types to be acceptable. In another 20\% of the  cases (5/25), the predicted types pertained to additional semantic facets of the entity; e.g., an entity labeled as a {\it writer} was predicted, also correctly, as a {\it politician}. Another entity labeled as a {\it writer} was assigned the type of a {\it Company}; this entity account belonged to an individual, who managed commercial activity of publishing books and other content online. Thus, in 60\% of the sampled examples with type discrepancy, the predicted types were semantically meaningful.

In the remaining cases, we found that the classifier was sometimes misled by the topical domain of the content posted by the account. For instance, a {\it website} about financial investments was predicted to be a {\it business person}, and a {\it business person} posting content similar to a blog was predicted to be a {\it writer}. While inaccurate, such mistakes (7/10) were topically meaningful. Only in 12\% (3/25) of the cases, we found the predictions to be blunt errors: mistaking a {\it musical artist} with a {\it basketball player} or a {\it politician}, and replacing a {\it company} with a {\it music band}. In all of those cases, we found that the tweets posted by those accounts were semantically irrelevant with respect to the account's semantic type and therefore misleading. 

In summary, we find that the majority of prediction gaps reflect the subjective and non-deterministic nature of semantics. In cases where the predicted labels differed from the types selected by the annotator, they were often semantically valid, or informative. We find the fact that as few as 12\% of the investigated errors were due to misleading or insufficient evidence to be encouraging. In particular, this means that the automatic prediction of the pool of unlabeled popular Twitter accounts provides a meaningful approximation of their types. 

\subsection{Mapping the semantics of the Twitter Sphere}

\begin{figure}[t]
\centering
\includegraphics[width=0.75\columnwidth]{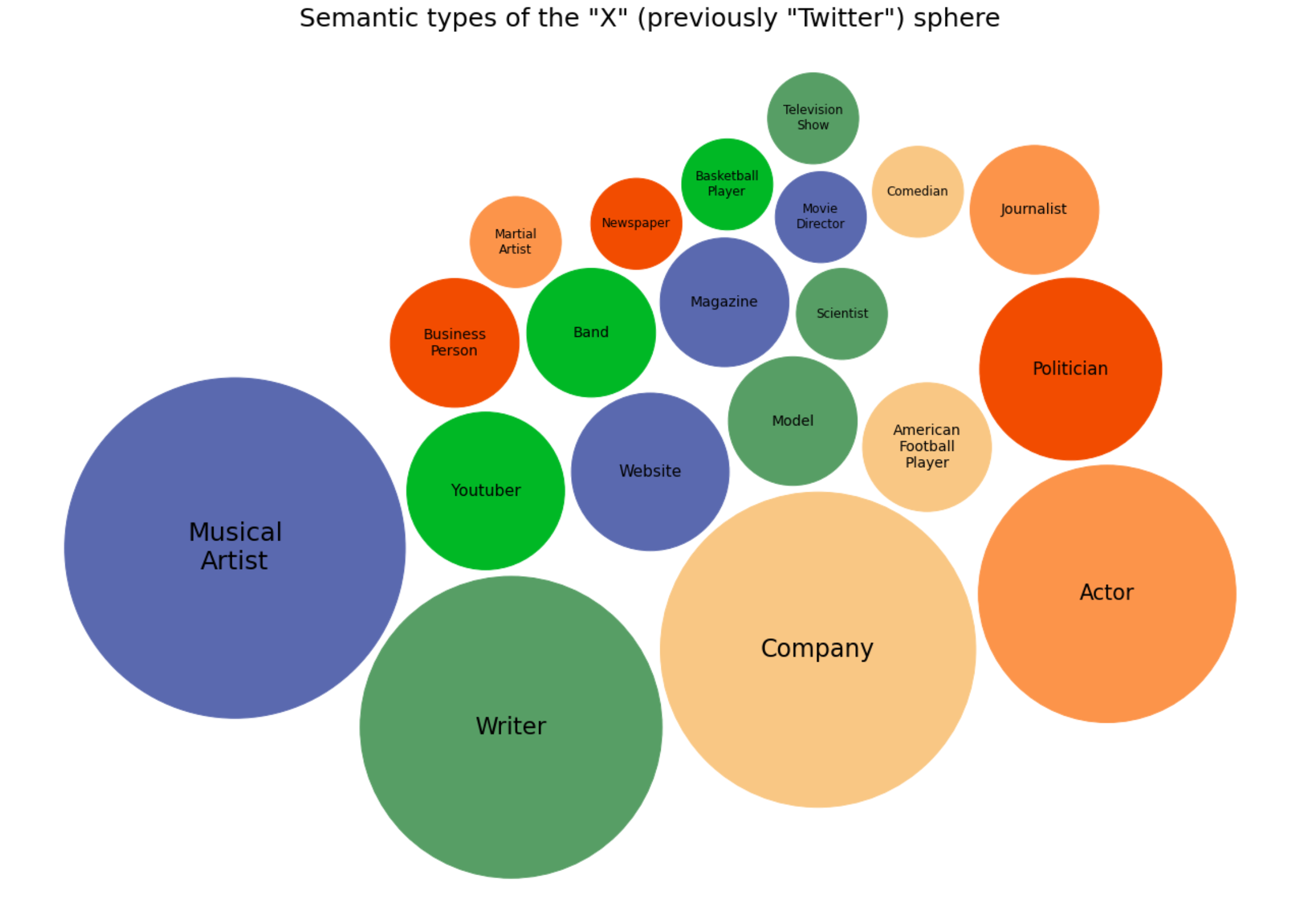}
\small
\caption{A visualization of the predicted distribution of semantic types among 200K popular Twitter accounts.}
\label{fig:final-dist}
\end{figure}

Finally, we apply our type classification model to the majority of unlabeled entities ($\sim$157K) within the social KB. By that, we address the research question stated in the beginning of this paper, "What are the semantic types of popular entity accounts on Twitter?" This large-scale process involved the processing of the historical tweets and network information of those accounts into user-level embeddings, and performing type classification given this evidence using our best-performing model. Overall, this large-scale classification step required roughly 10 hours of processing time using a standard Intel core i7 processor.

\begin{table}[t]
\small
\centering
\begin{tabular}{llrr}
Coarse & Fine & Ratio [\%] & Cumulative [\%]\\
\hline
Person	&	Musical Artist	&	0.14	&	0.14	\\
Org	&	Company	&	0.12	&	0.26	\\
Person	&	Writer	&	0.11	&	0.38	\\
Person	&	Actor	&	0.08	&	0.46	\\
Person	&	Politician	&	0.04	&	0.50	\\
Work	&	Website	&	0.03	&	0.53	\\
Person	&	Youtuber	&	0.03	&	0.56	\\
Person	&	Journalist	&	0.02	&	0.58	\\
Person	&	Model	&	0.02	&	0.60	\\
Person	&	American Football Player	&	0.02	&	0.63	\\
Org	&	Band	&	0.02	&	0.65	\\
Person	&	Business Person	&	0.02	&	0.66	\\
Work	&	Magazine	&	0.02	&	0.68	\\
Person	&	Scientist	&	0.01	&	0.70	\\
Person	&	Comedian	&	0.01	&	0.71	\\
Work	&	Newspaper	&	0.01	&	0.72	\\
Person	&	Basketball Player	&	0.01	&	0.73	\\
Person	&	Soccer Player	&	0.01	&	0.73	\\
Work	&	Television Show	&	0.01	&	0.74	\\
Person	&	Martial Artist	&	0.01	&	0.75	\\

\hline
\end{tabular}
\caption{The top 20 semantic types assigned to the entire 200K popular Twitter accounts in our social KB by alignment (43K) and classification (157K).}
\label{tab:final-dist}
\end{table}

The main types in the resulting distribution are detailed in Table~\ref{tab:final-dist}, as well as illustrated in Figure~\ref{fig:final-dist}. It is interesting to contrast this distribution, most of which has been obtained using automatic prediction, against the type distribution among the accounts aligned with DBpedia (Table~\ref{tab:dataset}). We observe that in the general population of popular Twitter accounts, there is a higher ratio of {\it Writer} (11\% vs. 4\%), {\it Musical artist} (14\% vs. 12\%), and {\it Website} (3\% vs. 2\%). Further, the categories of {\it Youtuber} (4\%), {\it Model} (2\%) and {\it Business person} (2\%) all measured a small fraction (1\% of less) within the aligned entities. Inversely, the proportion of {\it Company}, {\it Actor}, {\it Politician} and {\it Soccer player} within the whole KB is lower compared to the aligned accounts, measuring 12\% vs. 14\%,  8\% vs. 12\%, 4\% vs. 5\%, and 1\% vs. 2\%, respectively. We believe that compared to widely known entities that are included in factual KBs, many popular accounts in the social sphere correspond to content creators, bloggers. and interest groups, which are not covered by such resources. In particular, we hypothesize that the category of {\it Writer} that is prevalent within the social KB often pertains to bloggers or individuals engaged in online content creation. Similarly, following our error analysis, we believe that many accounts that were assigned the type of {\it Musical artist} correspond to accounts that present or discuss musical content online. Likewise, a portion of the accounts classified as {\it Basketball player} or {\it Company} merely discuss basketball or content of commercial orientation, respectively. Since our type labeling scheme matches the aligned accounts, the classifier would assign the most relevant type based on available evidence. Despite this limitation, this study is first to present an informative large-scale mapping of the semantic types and topics of popular Twitter accounts.

\section{Applications: Social entity similarity}
\label{sec:sim}

Our work presents a couple of main artifacts: (i) Coarse and fine-grained semantic types, assigned to each entity account in the social KB, and (ii) Content-based semantic entity embeddings, produced using the BERT model as finetuned on the semantic type prediction task. There is a variety of applications for which semantic entity types and embeddings are useful, including:
\begin{itemize}
    \item {\it Semantic text enrichment.} Researchers have shown that incorporating semantic entity embeddings alongside word embeddings can enhance text processing tasks. Similar to previous works~\cite{wiki2vec,ebertEMNLP2020}, social entity embeddings can provide useful context information about entities that are mentioned in text.
    \item {\it Entity retrieval.} In information retrieval, it is desired to rank entities by relevance to the query. For examples, researchers studied a scenario where entity mentions are first identified in the query (e.g., "What companies are {\it Microsoft}'s competitors"), and candidate entities are then  by their similarity to the query entity in the semantic embedding space~\cite{gerritseECIR2020}. Likewise, social entity embeddings may be utilized to incorporate a notion of semantic and social similarity in entity retrieval. 
    \item {\it Question answering.} This application aims to retrieve an entity of in response to a query~\cite{yavuzEMNLP16}. For example, the response to the question "Where is the Louvre located?" is "Paris". In  this setup, a semantic match of the answer type has to be verified. For example, questions that start with "where" require an entity of type {\it location} as answer. Likewise, fine-grained entity types may help address questions such as "Who is the Mayor of NYC?"~\cite{choi2018ultra}. Our entity embeddings and types may leverage question answering in general, and on social media in particular~\cite{tweetQAACL19}.
    \item {\it Entity recommendation.} Recommending entities of interest to users based on their network and other personal information on social media is a well studied task~\cite{hannonRecSys10,yuWSDM14,pritskerIUI17}. Given the network- and text-based embeddings of the entities, one may process entity recommendation of social and semantic flavor with respect to a user, as well as a topic. The semantic types assigned to the entities make it possible to restrict the recommended entities to a desired type. For example, entities of type {\it author} for which the content embeddings are similar to the embeddings of the topical words "culinary" and "foods" may help identify authors of books in this domain. 
\end{itemize}

In most of the applications mentioned above, assessing entity similarity is a key ingredient, where similarity is computed in terms of cosine similarity in the embedding space~\cite{yamadaEMNLP2020}. In this section, we demonstrate the utility of the entity embeddings presented in the work for the task of entity similarity assessment. 

\subsection{Computing social and semantic entity similarity}

The entity similarity assessment task is defined as follows. Given a query entity $e_q$, it is desired to rank other entities $e_j \in E$, where $E$ is a pool of candidate entities, by their similarity to that entity. Typically, all of the accounts for which compatible encodings exist are considered, where relevance to the query is computed using cosine similarity. 

Table~\ref{tab:example_rankings} showcases example similarity assessments, listing the topmost five similar entities per several query entities using either network-based SocialVec or the content-based semantic embeddings which we generated using the finetuned BERT in this work. The example queries include the entrepreneurs {\it Bill Gates} and {\it Elon Musk}, the {\it New York Times} magazine, and the company of {\it SpaceX}. As shown, each representation yields different results. The rankings produced by the textual encodings model a notion of semantic similarity, favoring entities of other business people, news outlets and aerospace companies per the given queries, respectively. Content-based ranking lacks however a necessary notion of social relevance. For example, Jeff Bezos is ranked lower with respect to Bill Gates than with Shahzad Rafati, the chairperson and CEO of a global media company, yet Rafati is not a business mogul like Gates or Bezos. Likewise, Courthouse News is found to be most similar to the New York Times; while both are news outlets, the match seems awry  as Courthouse news is a niche news service focusing on civil litigation. Computing similarity using the SocialVec embeddings encodes social relevancy and favors popular entities, yet it sometimes fails to model semantic similarity~\cite{lotan21}. For example, the most similar entity to Elon Musk in the SocialVec embedding spaceis the company that he has established, Tesla. Despite being topically relevant, the two entities are of different semantic types--({\it person} vs. {\it company}). Likewise, the entities that found to be similar to the {\it New York Times} magazine are semantically eclectic, including the accounts of {\it Google}, the singer {\it Alicia Keys}, and the comedian {\it Conan O'Brian}. This ranking suggests that users who follow the account of the NYT tend to be interested also in these other popular accounts, failing to capture semantic similarity. These trends are observed across all queries. 

\paragraph{A combined measure of social entity similarity.}

Ideally, a measure of entity similarity should reflect multiple facets of relevancy, including semantic, topical and social relevancy. In the first part of this work, we showed that the text- and network-based encodings were complementary for the purpose of predicting the semantic types of entities. Next, we show that combining the two types of entity embeddings is preferable also in assessing social entity similarity.

The social and text-based embedding spaces are incompatible. Our attempts to compute cosine similarity between the concatenated vectors, or to rank entities by a weighted average of their similarity scores using each method, gave poor results. Rather, we found it beneficial to unify the information sources via {\it re-ranking}~\cite{minkovTOIS11}. Let a ranked list of entities be first produced using method $f_1$. It is assumed that the top-$k$ entities in the ranked list include some responses of high quality. As a second step, additional information is incorporated using another method $f_2$ to score and re-rank the retrieved pool of $k$ candidates, producing a final ranking.  

Here, we assign $f1$ to be the text-based similarity, where it is expected that the top ranked entities show semantic (type) similarity to the query. In our evaluation, we set $k=50$, based on manual inspection of candidate relevancy within the the lists of responses generated for other random queries. We then compute similarity scores for those $k$ candidates using network-based SocialVec similarity as the re-ranking method, $f_2$. In this manner, semantic similarity is prioritized in the candidate selection phase, where reranking presents the results in a socially sensible order. 

\subsection{Entity similarity: example results}

\begin{table}[t]
\small
\centering
\begin{tabular}{lllll}
\hline	
Query& Content (BERT) & Network (SV) & BERT$\circ$SV & Wikipedia2Vec \\
\hline
Bill Gates&John Doerr&Richard Branson&Richard Branson & Melinda Gates\\
&Sean Parker&FBI&Jeff Bezos & Steve Ballmer\\
&Richard Branson&TED Talks&Eric Schmidt & Microsoft \\
&Shahrzad Rafati&The New York Times&Tim Cook & Warren Buffett\\
&Jeff Bezos&Google&Melinda Gates & John Doerr\\
\hline
Elon Musk&Scott Manley&Tesla&Chris Dixon & SpaceX \\
&Sophie Alpert&Stephen Colbert&Biz Stone & Neuralink \\
&Rene Ritchie&Bill Gates&Alexis Ohanian & Brian Chesky\\
&Tom Warren&Reddit&Tobi Lutke & Blue Origin\\
&Lisa Randall&Bloomberg&Chelsea E. Manning & Kimbal Musk\\
\hline
New York Times&Courthouse News&Google&HuffPost & The Washington Post \\
&NPR&Bill Gates&The Associated Press & Los Angeles Times\\
&New York Times Opinion&Alicia Keys&NPR--National Public Radio& The Wall Street Journal\\
&BuzzFeed News&Conan O'Brien&The Washington Post & Michiko Kakutani\\
&The Real News&Jay-Z&NBC News & Frank Rich \\
\hline
SpaceX & ULA (United Launch Alliance) & Tesla&NASA Social & Blue Origin \\
&Blue Origin&NASA InSight&Intl. Space Station & United Launch Alliance\\
&Arianespace&Jeff Bezos&NASA Technology & Masten Space Systems\\
&Aerojet Rocketdyne&Simone Giertz&Reid Wiseman & Ariane Space\\
&Bigelow Aerospace&NASA Social&Lockheed Martin & Elon Musk\\
\hline
\end{tabular}
\caption{Examples of query entities, and the most similar entities as computed by inter-entity cosine similarity using several embedding schemes: the finetuned text-based BERT entity embeddings, which represent tweet history; the network-based SocialVec entity embeddings (V); and, their combination, having the top 50 candidates retrieved using the content BERT embeddings sorted by the network-based SocialVec similarity. The table presents also the top entities ranked using the embeddings of Wikipedia2Vec, derived from Wikipedia. The results are presented in descending order of the similarity scores.}
\label{tab:example_rankings}
\end{table}

Table~\ref{tab:example_rankings} presents the results of combining the two metrics via reranking (`BERT$\circ$ SV'). As shown, the top response to the query of Bill Gates is Richard Branson, where both entities are globally-known entrepreneurs. Notably, Branson was not included within the top-5 results of neither content- or network-based similarity. Thsi entity was included however in the top-50 candidates based on content similarity, out of which it was assigned the highest social network-based similarity. Likewise, Jeff Bezos was promoted from the fifth to the second rank using reranking. Due to conditioning on content similarity, the reranking approach presents only {\it persons} as the most similar entities to this query. A similar effect is demonstrated for the other queries. The most similar entity to Elon Musk using reranking is Chris Dixon, a co-founder of Twitter, which has been acquired by Musk. The top result for the query of New York Time magazine is HuffPost, which is another prominent magazine with a similar social orientation. And, the top result for the SpaceX company is an account of NASA, a federal U.S. agency responsible for the civil space program. In all of these cases, multi-facet similarity is achieved, yielding responses that are sensible with respect to semantics and  social knowledge, as reflected by network patterns on Twitter. 

\paragraph{A comparison with factual entity similarity.} 
Our entity encoding schemes are social in that they were derived directly from social media. For comparison, Table~\ref{tab:example_rankings} presents the top entity rankings produced using Wikipedia2Vec, a popular entity embedding scheme learned from Wikipedia~\cite{yamadaACL16,yamadaEMNLP2020}. Notably, the factual source of Wikipedia has a different scope from social media, covering many abstract concepts, historical entities, events, locations, etc. In order to allow a meaningful comparison, we limited the pool of candidates ranked using the Wikipedia2Vec method to entities that are included in our social KB. 

As shown, our composite social entity similarity measure and Wikipedia2Vec deliver different flavors of similarity. Wikipedia2Vec encodes relational associations between entities, learned from the hyperlinked structure of Wikipedia. For example, given the query of Bill Gates, Wikipedia2Vec retrieves entities that are related to Gates: Melinda Gates, his former partner; the company which he founded Microsoft, and its former CEO Ballmer;  Warren Buffet who served as a member of the Bill and Melinda Gates Foundation board; and, John Doerr, the author of a best selling book about successful organizations such as the Gates foundation. Our composite social measure rather returns a list of other world-known entrepreneurs, like Richard Branson (a co-founder the Virgin group), Jeff Bezos (Amazon), Eric Schmidt (a former CEO of Google), Tim Cook (the CEO of Apple), as well as Melinda Gates. As another example, in response to the query of NYT magazine, the top results using Wikipedia2Vec include also columnists of the NYT (Kakutani and Rich), whereas the social measure is more focused on magazines and news sources, i.e., relevant entities of a similar semantic type to the query. 

Our evaluation on additional queries showed similar trends. While anecdotal in nature, this evaluation suggests that:
\begin{enumerate}
    \item SocialVec and and the content-based entity embeddings highlight complementary social and semantic aspects of entity similarity. Reranking is an effective method for combining the two measures, serving to identify entities of high semantic and social relevance to a query.  
    \item Both of the social entity embedding schemes offer a different, and hence complementary, perspectives of entity similarity compared with relational similarity that is derived from factual sources.
    \item Our social entity embeddings form a valuable source for computing entity similarity by downstream applications for many popular entity accounts, which are active on social media, but are not covered by factual KBs.
\end{enumerate}

\section{Conclusion}
\label{sec:conclusions}

In this work, we explored the semantics of a  social knowledge base that includes a large number of popular Twitter entities. Since there is no explicit semantic annotations on social media, we outlined a framework for inferring entity semantics automatically. Aligning Twitter accounts to the factual KBs of Wikidata and DBpedia, we extracted and processed 136 fine grained semantic types, which we attributed to the aligned entities according to those sources. Using the resulting dataset of roughly 34K labeled entities, we then learned type prediction models using public evidence that is available on social media. Our experiments indicated that content-based embeddings, which represent the account's post history as encoded by a model of BERT that has been finetuned using our dataset encode informative entity semantics. We further found that network-based embeddings provide complementary evidence about entity semantics. 

\paragraph{Main contributions and results.} Overall, we obtained weighted F1 performance of nearly 68\% on the task of fine-grained type prediction using the labeled dataset. Henceforth, we applied our model to the remaining entities within our KB of Twitter entities to predict their types. Thus, we have enriched the social knowledge base of 200K entities with semantic entity embeddings and type assignment for each entity. Finally, we have demonstrated the utility of the social entity embeddings generated and evaluated in this work on the task of entity similarity. Our evaluation on example instances has shown that the composition of network- and content-based embedding similarity yields high-quality similarity assessments. Importantly, the social entity similarity complements existing factual entity embeddings, as it encodes social perception and trends rather than relational facts, and in that is applies to a large number of entities which solely reside on the social media sphere.

\paragraph{Limitations and future work.} A main limitation of our framework is the possibility of a distribution shift between the Twitter entities that comprise our dataset of aligned labeled entities, using which our prediction models were learned and tuned, and the general population of Twitter entities in the social KB. A manual analysis which we reported in this work indicated on some faulty prediction due to scarce or noisy evidence that is available for some of the accounts. Schema-wise, there are type categories that are unique or prevalent on social media, which are not well represented in the labeled dataset, e.g., {\it content creators}. Future work may address this gap by manual labeling, which is costly, or perhaps using another form of distant supervision, to learn type prediction models from a random sample of Twitter entities. Nevertheless, we note that despite this limitation, we believe that the semantic entity embeddings which we learned in this work can support a variety of downstream applications. 

There are many exciting future directions of this research. As discussed above, the social entity representations may leverage a variety of applications that involve entity linking, recommendation and representation. In particular, we are interested in integrating social knowledge in conversation systems. Given a conversation topic, one may retrieve relevant entities of social and topical relevance using our entity embeddings, composing engaging system responses that involve those entities. Furthermore, assuming that the system users may be represented in terms of popular entities that they follow on social media~\cite{lotan21}, then relevant entity selection may be personalized to match the users' preferences. Likewise, entity disambiguation and linking may be personalized by means of social entity similarity.

\bibliographystyle{ACM-Reference-Format}
\bibliography{main}

\end{document}